\documentclass[a4paper,11pt]{article}

\usepackage{arxiv}
\usepackage[spanish,english]{babel}

\usepackage[utf8]{inputenc} 
\usepackage[T1]{fontenc}    
\usepackage[hidelinks,linktoc=all,colorlinks=true,linkcolor=blue]{hyperref}
\usepackage{url}            
\usepackage{booktabs}       
\usepackage{amsfonts}       
\usepackage{nicefrac}       
\usepackage{microtype}      
\usepackage{lipsum}         
\usepackage{graphicx}
\usepackage[numbers]{natbib}
\usepackage{doi}
\usepackage{xspace}
\usepackage{bbm}

\usepackage{comment}

\usepackage{longtable}

\usepackage{marvosym}

\usepackage{graphicx} 
\usepackage{adjustbox} 

\usepackage[inline]{fixme}

\usepackage{tocloft}

\usepackage{graphicx}
\graphicspath{{../Pictures/}}

\usepackage{setspace}
\usepackage{titlesec}
\titleformat{\chapter}[hang]{\Huge\bfseries}{\thechapter}{1em}{\Huge\bfseries}

\usepackage[detect-weight]{siunitx}
\usepackage{lineno}
\usepackage{fixme}
\fxsetup{
    status=draft,
    author=,
    layout=inline,
    theme=color
}

\usepackage{datetime}

\usepackage{rotating}
\usepackage{longtable}
\usepackage{setspace}

\usepackage{amsmath}

\usepackage{mdwlist}
\makecompactlist{compactitemize}{itemize}

\usepackage{algorithm}
\usepackage{algorithmic}

\newif\ifpdf
\ifx\pdfoutput\undefined
  \pdffalse
\else
  \pdfoutput=1
  \pdftrue
\fi

\ifpdf
\usepackage{pdflscape}
\else
\usepackage{lscape}
\fi

\renewcommand{\undertitle}{}
\renewcommand{\headeright}{}

\hypersetup{
pdftitle={Sports center customer segmentation: a case study},
pdfsubject={customer segmentation, clustering, },
pdfauthor={Juan Soto, Ramón Carmenaty, Miguel Lastra, Juan Manuel Fernández-Luna, José M. Benítez},
pdfkeywords={Sports Centers, Customer Segmentation, k-means, Similarity Optimization},
pdfcreationdate={20240429}
}

\ifpdf\fi

\usepackage{authblk}

\title{Sports center customer segmentation: a case study}

\author[1]{Juan Soto}
\author[1]{Ramón Carmenaty}
\author[2]{Miguel Lastra}
\author[1]{Juan Manuel Fernández-Luna}
\author[1]{José M. Benítez}

\affil[1]{Computer Science and Artificial Intelligence Dept.\\
DiCITS Lab, iMUDS\\
Universidad de Granada}

\affil[2]{Software Engineering Dept.\\
DiCITS Lab, iMUDS\\
Universidad de Granada}

\date{20 May 2024}

\newcommand{\kmeans}{$k$-means\xspace}
\newcommand{\etal}{\emph{et al.{}}\xspace}

\begin{document}

\maketitle

\begin{abstract}
Customer segmentation is a fundamental process to develop effective marketing strategies, personalize customer experience and boost their retention and loyalty.
This problem has been widely addressed in the scientific literature, yet no definitive solution for every case is available.
A specific case study characterized by several individualizing features is thoroughly analyzed and discussed in this paper.
Because of the case properties a robust and innovative approach to both data handling and analytical processes is required.
The study led to a sound proposal for customer segmentation.
The highlights of the proposal include a convenient data partition to decompose the problem, an adaptive distance function definition and its optimization through genetic algorithms.
These comprehensive data handling strategies not only enhance the dataset reliability for segmentation analysis but also support the operational efficiency and marketing strategies of sports centers, ultimately improving the customer experience.
\end{abstract}


\section{Introduction}




The ability to segment and understand customer needs and behaviors is fundamental for developing effective marketing strategies, personalizing the customer experience, and ultimately enhancing retention and loyalty.
Customer clustering emerges as a powerful analytical technique that allows businesses to group customers into homogeneous segments, seeking similarity among the characteristics of customers within the same segment.
These segments are then used to develop specific marketing strategies that address the needs and interests of each group.


While general proposals can be found in the literature, the wide variability of areas make difficult to design a general methodology for customer segmentation.
That makes it worthy to deeply study particular cases whose solutions can later be generalized to similar cases.
In this work a real world problem with particular considerations is addressed.
The main objective was to optimize business operations and customer satisfaction in sports centers using Artificial Intelligence techniques.
To achieve this, customer clustering plays a crucial role by enabling effective segmentation of the user data base.
Customer data usually encompasses training frequency, use of sports facilities, or other types of data such as demographics.

The main specific challenges encountered in the case under study are as follows: a very large customer database with records of over 3 million clients and many missing pieces of data in those records.
As numerous database entries were impacted by the absence of a value for certain variables and at the same time many different variables were affected, dropping samples with missing values had to be discarded as it would have greatly diminished the amount of available data.
Furthermore, since the missing values issue affected not only numerous samples but also numerous variables, dropping columns could neither be considered an acceptable solution.
This situation was tackled by implementing a data partition scheme where samples within in the same region had similar characteristics in terms of data availability.

The clustering process itself was designed using a 2-step process using a combination of DBSCAN \cite{ester1996density} to get an initial cluster set without having to manually set an specific number of clusters and \kmeans \cite{lloyd1982least} to refine the clusters obtained from the previous step.
As the clustering process is highly impacted by the distance function employed, a specialized customer dissimilarity function has been created.
First, by applying data non-uniform transformations that takes into account the specific data distribution and to minimize the effect of outliers.
Second, approximating the variable importance through the predictive models built to solve artificial designed related supervised problems and incorporating them through weights in the distance function. And, third, optimizing the weight values through a genetic algorithm.
This algorithm solved the optimization problem which involved finding the optimal weight for each variable in the distance function to maximize the quality of the clusters generated by \kmeans.
Thus, the genetic algorithm was used to find the optimal relative importance of each variable when computing the distance between a pair of samples with the clustering process in mind.

The designed methodology has proved to be very effective. Its application to the study case has resulted in a set of groups found to be very relevant for the expected marketing goals.
Anyhow, it is worth noting that the success of customer clustering largely depends on the quality of data and a deep understanding of the business domain.
Additionally, the effective implementation of strategies derived from clustering requires ongoing commitment to data collection, iterative analysis, and adaptation as customer needs and preferences evolve.

\section{State of the art}

A bibliographic search was conducted on the scientific literature regarding customer segmentation methods in various fields, particularly in the realm of physical activity, and extended to other domains.
We discuss the most relevant works in this section. 

\subsection{General Customer Analysis}

The scientific literature in the area of customer analysis and profiling is extensive and varied.
One of the main characteristics of customer analysis is the high dimensionality of the problem.
It is common to have a large number of variables associated with these customers, implying that the process of discovering groups must deal with this high dimensionality and often with sparse information in that space.

Aggarwal \etal \cite{Aggarwal199961} address this problem considering that in different groups, the appropriate subset of features to use may not be the same, and propose a projected clustering method where different sets of features specific to each cluster can be used. This work is extended to large databases in \cite{Aggarwal2002210}. Additionally, besides high dimensionality, the issue of qualitative information, not just quantitative, is prevalent. Thus, Gil David and Amir Averbuch presented SpectralCAT \cite{David2012416}, the diversity of data types using spectral clustering through projections into lower-dimensional spaces is also addressed.

Mets \etal \cite{Mets20162250}, in the field of data obtained from electrical networks, present a method based on wavelet transformation to reduce dimensionality. This aspect enables the scalability of the presented method. Finally, in the context of customer data from telecommunications companies, Alkhayrat \etal \cite{Alkhayrat2020} present a dimensionality reduction method based on Principal Component Analysis (PCA). In this work neural networks are used to project to a lower-dimensional space combined with the \kmeans technique.

In many cases, customer segmentation is primarily driven by economic interest, where the value of the customer to the company is a central issue. In the field of wireless telecommunications, Hwang \etal \cite{Hwang2004181} and Kim \etal \cite{Kim2006101} propose methods based on estimating the long-term value of the customer, Customer Lifetime Value (LTV). This customer value, compared to the cost of maintaining them as such, allows determining the effort that should be applied to maintain their loyalty. In relation to these aspects, Lee \etal \cite{Lee2005145} present a segmentation method to extract the most valuable customers using an estimation of their satisfaction level. The presented technique utilizes multi-agents, business intelligence tools, neural networks (self-organizing maps), the C4.5 algorithm, and is applied to data from a company in the automotive segment.

In the field of financial data it is common to use customer valuation techniques (scoring) to establish associated risk levels. Hsieh \etal \cite{Hsieh2004623} present a clustering method at various levels of expected customer gain from a financial entity. This work also highlights the difficulty associated with the inherent multidimensionality of the handled data.

Given the complexity of the task at hand, some authors have resorted to hybrid approaches to address the problem using different techniques. Kuo \etal \cite{Kuo2016116} present a metaheuristic-based clustering method. In a first phase, co-association and co-occurrence matrices are created on which PCA is applied to reduce dimensionality. Subsequently, different techniques based on metaheuristics are applied: a genetic algorithm, one based on particle swarms, and an optimization method based on artificial bee colonies. In \cite{Li2021}, \kmeans is combined with particle swarm optimization applying adaptive learning to reduce the chances of falling into local minima. Finally, as a way to address the fact that finding a segmentation algorithm that can be successfully applied to datasets of very different nature is highly complicated, in \cite{Kuo2018299}, a hybrid method based on a metaheuristic method and Fuzzy C-Means is proposed. An evolutionary method is used to obtain an initial approximation of the centroids of possible clusters, which are later used as a starting point by the fuzzy clustering method. There are other examples of applying methods based on fuzzy logic, such as that presented by Crespo \etal \cite{Crespo2005267}, applied to user grouping using traffic management-related information.

In the context of searching for the optimal partition of clients into disjoint groups based on information about their transactions, the use of direct clustering techniques is proposed by \cite{Jiang2009305} applying techniques that attempt to approximate the optimal partitioning in a computationally feasible manner since it is an NP-hard problem. This is done using techniques that, while suboptimal, produce results close enough to the ideal solution.

Finally, aimed at improving recommendation systems, in \cite{Liu20093505}, a hybrid technique is presented that combines information from the sequence of customers' purchases with KNN-CF to take into account that the purchasing profile behavior of customers can change over time. Thus, KNN-CF is exclusively applied to the most recent purchase sequences and not to the complete history of purchase sequences available.

\subsection{Customer Analysis in the Sports Activity Sector}


Customer or user segmentation has been widely studied in many areas. Some examples are  electricity consumption, financial entities (banks, credit institutions, etc.), telecommunications, and online sales.
Along the last years, a growing sector is services oriented towards improving or maintaining the physical condition of people: gyms, sports centers, or fitness centers.
However, to the best of our knowledge, in this specific field the reported research is rather scarce.

In \cite{Jang20181043}, a study was conducted on the reasons why people decide to become customers of such centers.
It was concluded that the membership fee, the environment, accessibility, and finally the quality of the equipment were the main factors.
It was also studied whether the time spent as a customer influenced altering the relative order of these factors.

In the sports field in general, there are works on segmentation of sports enthusiasts regarding the purchase of season tickets \cite{Finch2022110}, segmentation of customers regarding the purchase of equipment for sports practice \cite{Rogic202171}, and specifically using information about movements within the establishment selling such equipment \cite{Carbajal20211}.
Finally, using data from sports monitoring devices, the creation of user groups has also been studied \cite{Paweloszek20214751}.

\section{Sports centers: a case study}

%
%


As the fitness industry grows and diversifies, understanding the diverse needs and behaviors of sports center patrons becomes crucial.

A leading company in the sector contacted our research group requesting assistance to help them generate additional value through enhanced personalization of services. This turned into a customer segmentation problem.
The solution should derive into actionable insights. 

A customer segmentation process is targeted to identify relevant and cohesive client segments.
This not only enhances the operational efficiency of sports centers but also significantly improves the overall customer satisfaction.
Further enhancing the operational implications, this analysis directly assists gym managers in carrying out their tasks more effectively.
Providing managers with segmented data and customer insights enables them to make informed decisions on service modifications, marketing adjustments, and customer engagement strategies.
This targeted support helps managers optimize resource allocation and improve the overall customer experience at their facilities.

The company provided us with a representative dataset of their customer data.
It was composed of different features of customers from a large set of sports centers.
Actually, it enlisted close to 3.5M customers from multiple sports centers, incorporating various chains and locations across different countries, spanning for a period of over two years.
Company experts selected a dataset of 44 potentially relevant variables.
These included demographic information ---e.g. sex, age, country, and province---, physical condition, customer stated targets, activity preferences, booking habits, and routines ---e.g. frequencies of attendance, preferred activities, training tasks.
Altogether, they offer insights into each member’s relationship with the sports center, behavioral patterns and preferences.

This diversity in data sources brings with it a unique set of challenges, particularly in terms of data capture and quality.
Technologies and methods for gathering data vary widely among sports centers; while some employ advanced digital tracking systems and integrated mobile applications, others might rely on less sophisticated or manual data entry systems.
This variation significantly affects the consistency and reliability of the data collected, introducing issues such as incomplete records, inconsistent entries, and varying timeliness of data availability.
To address these discrepancies, crucial steps have to be taken to ensure that the analysis could lead to reliable results.


The study aims to classify customers into meaningful segments by closely examining their engagement patterns and preferences, which in turn helps in offering more personalized services. 
A segmentation study not only highlights the typical profiles of individuals who frequent these facilities but also addresses specific challenges and nuances associated with the dataset used.





\section{Customer segmentation procedure}
\label{sec:segmentation-procedure}

The trek for an effective solution to our case study led to the definition of a complete workflow for customer segmentation.
In this section, we detail the algorithm process that we conceived and carried out to fulfill the task.

Since no specific target guide is provided, the process is altogether unsupervised and a clustering algorithm is the core of it. 
However, a number of previous steps are essential.
While all the technologies and algorithms used are already known, the innovation in the process lies in the original combination and integration, as well as the adaptation to the peculiarities of the specific case. 
We begin by presenting the complete process, that is outlined as shown in Algorithm~\ref{alg:customer-segmentation}.

\begin{algorithm}
\caption{\label{alg:customer-segmentation}
Customer segmentation procedure}
\begin{algorithmic}[1]
\STATE Exploratory data analysis.
\STATE Data space (and dataset) partitioning.
\STATE Data preprocessing and transformation.
\STATE Determining the number of clusters.
\STATE Distance function definition.
\STATE Distance function optimization.
\STATE Clustering.

\end{algorithmic}
\end{algorithm}

The process structure is fairly straighforward.
However, the dataset complexity, volume, and data features require a robust and innovative approach to both data handling and analytical processes.
In particular, the approach to definition and optimization of the distance function has proved to be very effective and, to the best of our knowledge has not been reported yet.

The steps of the procedure are detailed along this section. The whole procesing pipeline has been implemented in the Python programming language \cite{Python} using a number of libraries from its ecosystem. The most relevant are:  \texttt{Pandas} \cite{reback2020pandas} was essential for data analysis and manipulation, offering tools to clean, filter, perform statistical calculations, and handle null values through its Series and DataFrame structures. \texttt{Scikit-learn} (sklearn) \cite{pedregosa2011scikit} that provides efficient machine learning tools for classification, regression, clustering, and model selection. And, finally, \texttt{DEAP} (Distributed Evolutionary Algorithms in Python) \cite{DEAP_JMLR2012}, that implements evolutionary algorithms for optimization tasks and pattern discovery.

%
%
%

\subsection{Exploratory data analysis}
\label{sec:EDA}

As it is customary in data science processes, an exploratory data analysis (EDA) was conducted.
This preliminary analysis was crucial to gain initial insights in data quality, representativity and first patterns.
In special, it reveals the extent of missing data and the presence of outliers, which are key factors that could potentially skew the analysis and lead to unreliable segmentation results.

The EDA revealed severe data quality problems, notably a large amount of missing values in several relevant variables.
For instance, for some of the most relevant variables their percentage of missing values were too high: 57.21\% for physical access count, 30.78\% for total app access count, and 62.70\% for the number of total reserved activities. Similarly, booked activity reservations were absent in 63.42\% of cases, assigned trainings in 55.54\%, and validated trainings in 71.26\%. In addition, personal data variables like the fitness target and weight were highly unavailable with absence rates of 55.54\% and 68.08\%, respectively.

Additionally, another relevant outcome of the EDA process was outlier analysis.
Variables with outliers included age, which had an outlier percentage of 1.00\%, and total access count, with outliers constituting 9\% of its data.
Total app accesses and total reserved activities showed even higher outlier percentages: at 10\% and 11\%, respectively.
In fitness-specific metrics, assigned trainings and validated trainings had outliers in 4\% and 1\% of the data, while weight and body mass index (BMI) had unusually high outlier percentages at 1.98\% and 1.14\%, respectively.

To mitigate the identified challenges of data quality several measures were adopted.
To address the problems with missing data, we employed two primary strategies: imputation of missing values and using these incomplete variables to segregate the population before performing the desired segmentation.
This dual approach allowed us to maintain the integrity and usefulness of the dataset while adapting to the specific characteristics and inconsistencies present within the data.

Outliers, particularly those at the upper end of data ranges, present another significant challenge.
To objectively determine atypical values, we followed the established statistical convention of flagging any data points that lie beyond three standard deviations $(3\sigma)$ from the mean.
This method establishes a criterion to distinguish between normal data variance and true anomalies.
Additionally, when outliers are suspected to result from erroneous data entries, such as implausibly reported ages, we implement early detection filters.
These filters enable prompt correction, preserving the dataset’s accuracy.
We also examine outlier patterns that might be specific to certain geographic locations or centers, allowing for targeted remediation efforts that further refine our data quality.

Another significant piece of knowledge is the relevance of variables.
Since the dataset is composed of a high number of variables, not all of them were of the same importance. 
No specific target was provided, nor regular approaches to feature selection or feature ranking could not be applied.
Thus, we created \emph{artificial} machine learning problems to assist in the exploration and establish the correlations between variables.
These problems consist in predicting one of the variable out of a subset of the others.
Then predictive models for each problem were built.
To create these models, we employed Random Forests (RF) \cite{breiman2001random} and Extreme Gradient Boosting (XGBoost) \cite{chen2016xgboost}.
Both of them compute a measure of variable importance within the model built, and thus, the problem being solved.
The importance provided by the models is a ---partial--- view of the \emph{overall} variable importance.
By conveniently combining the importance of various models an \emph{aggregative} importance can be estimated for each variable. 
The results of these analyses directly influenced our decision-making on which variables to retain for in-depth analysis.

\subsection{Data space partitioning}
\label{sec:dataset-partition}

The EDA had revealed high rates of missing values in many variables.
Common approaches to tackle with this issue include data imputation, eliminating instances with missing data, or completely removing the affected variables.
Imputation, however, was complicated due to the low density of data for many variables, making approximation models ineffective.
Conversely, eliminating complete instances for each variable with missing data would result in an overly drastic reduction of the dataset.

We opted for an intermediate solution that did not involve the complete elimination of variables.
Given the dataset size and to maintain representativeness, we divided the data space into regions according to variable availability.
When applying this idea each variable gives rise to two subsets: one comprising instances that have values for the variable and another comprising those that do not.
Thus two zones for each variable are created and, when combined across several variables, the number of regions multiplies.

However, this method cannot be applied indiscriminately, as the number of regions grows exponentially with the number of variables used.
It was necessary to restrict this division by selecting only variables of high relevance and with a high number of missing values.
This strategy allowed us to maintain a balance between data integrity and the comprehensiveness needed for robust analysis.

\begin{figure}[h]
    \begin{center}
    \adjustimage{max size={1.0\linewidth}{0.9\paperheight}}{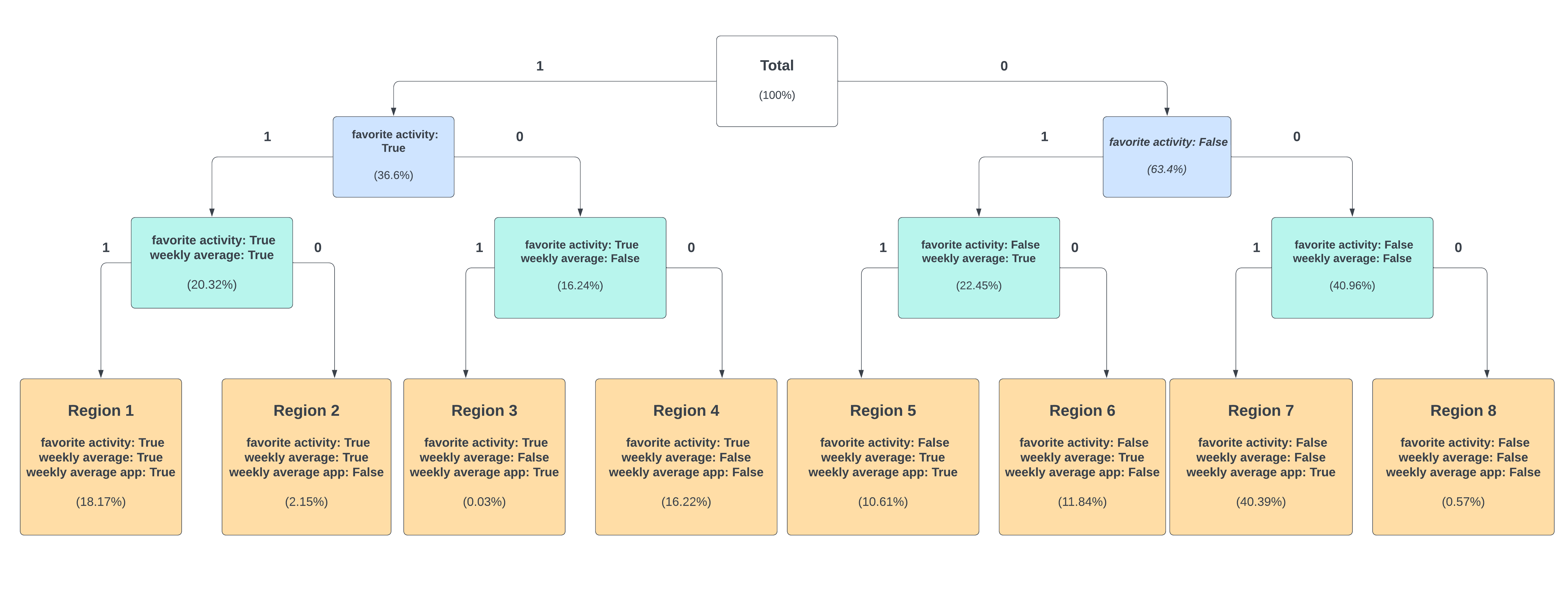}
    \end{center}
    \caption{Dataset division in regions}
    \label{fig:regions_division}
\end{figure}

A direct effect of this partitioning of the data space into regions is that the segmentation process ceases to be a global process run on the whole dataset
Instead a decomposition of the problem is done.
Now the initial segmentation is divided into a larger number of smaller, independent segmentation problems.
After careful analysis, the set of variables selected for the data space partition were: favorite activity, average number of accesses and average number of app accesses.
These variables were chosen not only for their relevance but also because they maintained a balance between completeness and informational value.
They induce a partition into eight regions of the data space, as depicted in Fig.~\ref{fig:regions_division}.
It is natural to describe plot as a binary tree, with eight leaves, each representing a different region.
The paths from the root node describe which of the selected variables are considered in the region (branch labeled with ``1'') or not (branch labeled with ``0'').
Each of the leave nodes includes the combination of dividing variables considered as well as the percentage of instances included.
Of course, the region sizes are hardly expected to be balanced. 

From this step on, each of the segmentation procedure is applied independently in each region.
The final segmentation is composed of the union of the clusters obtained in each partition.

\subsection{Data preprocessing and transformation}
\label{sec:preprocessing}

The core step for segmentation is a clustering process.
The clustering requires evaluating the similarity ---or distance--- between pairs of instances.
It is commonly computed through a distance function.
Since the dataset is composed of features of different nature, their representation is based on different data types, e.g.~some are discrete while other are real number-valued.
Their reference sets are not the same, in particular, they have different cardinals or interval lengths.
Notwithstanding, all of these variables are components of the vector that represents each instance of the dataset.
Thus the transformations applied depend on the type of data to be treated.
In particular, the treatment of continuous variables and categorical variables are different.

For continuous variables, we employed a non-linear normalization transformation, specifically tailored to accommodate the unique distribution of each variable.
This approach scales the values from the original domain, $[\min, \max]$ to the interval $[0, 1]$.
However, rather than applying a linear scale, our normalization considers the density of data within different value ranges, emphasizing more common values and diminishing the influence of outliers.
During the exploratory data analysis phase, we identified specific intervals with higher data density.
This information is used to truncate or adjust extreme values to the boundaries of these intervals, effectively reducing their impact on the analysis.

The treatment of categorical variables involves converting the categories into a numerical format so that the discrete variables can be mixed with real-valued variables. 
For this purpose, we utilized label encoding, a common technique where each unique category within a variable is assigned a distinct value ---an integer, that is subsequently embedded into $[0, 1]$.


\subsection{Determining the number of clusters}
\label{sec:determining-number-clusters}

Selecting the optimal number of clusters is a crucial step in the clustering process, directly impacting the effectiveness and interpretability of the segmentation.
For this purpose, we initially applied the DBSCAN (Density-Based Spatial Clustering of Applications with Noise) algorithm \cite{ester1996density}.
DBSCAN  does not require the number of clusters to be specified a priori and can identify outliers as noise allowing for a first approximation of the appropriate number of clusters.


However, we observed that sometimes the number of clusters produced by DBSCAN did yield clusters with good separation.
Thus, for those cases, we use DBSCAN computed number of clusters as a starting point and then explored its neighborhood.
Each potential $k$ was evaluated based on the clustering results it produced.
The evaluation metrics included intra-cluster cohesion and inter-cluster separation, essentially measuring how compact each cluster is and how distinct they are from each other.
The chosen value of $k$ for each region was the one that resulted in the most defined and separated clusters.
This criterion ensures that the clusters are not only statistically valid but also meaningful and interpretable from a business perspective.


\subsection{Distance definition}
\label{sec:distance-definition}

In the \kmeans clustering algorithm, the choice of distance function or metric is pivotal in defining the similarity between data points and determining the placement of centroids.
The most commonly used metric in \kmeans is Euclidean distance ---actually, squared Euclidian distance---, which treats all variables within an instance (vector components) with equal importance.
This approach is generally reasonable, especially after each variable has been normalized to ensure all values are on the same scale.
However, in our customer segmentation problem some variables are more influential than others.
This was highlighted in the EDA, where the contribution level of some variables was found to be greater than others.
To accommodate this disparity in relevance, we modify the Euclidean distance formula by incorporating weights into its definition, so that each weight represents the importance of the correspondent variable.

These weights allow the distance measure to reflect the varying significance of different variables more accurately.
As a starting point, the relevance measures obtained from predictive models in the EDA were considered. 
The cohesion and separation of clusters as measured by the clustering method objective were not good enough. 
So we decided to seek better weight values tailored to the specific needs of the clustering, and thus dependent on the clustering process objectives. The immediate question is  how to find the optimal weights that yield the best distance function between points to represent customer similarities effectively? These optimal weights can be calculated using an optimization algorithm.

\subsection{Genetic algorithms for distance definition optimization}
\label{sec:ga-distance-optimization}

The task of determining the optimal weight vector for the distance function in our clustering model is notably complex due to the variety of factors influencing customer behavior.
To address this challenge, we considered a robust optimization technique: genetic algorithms \cite{goldberg1989genetic}.
These algorithms are renowned for their robustness and effectiveness across a myriad of applications, making them an excellent choice for the distance function optimization.

The genetic algorithm employed mimics the process of natural selection.
It starts with an initial population of solutions. 
Every solution to our problem is represented as a vector where each component corresponds to one of the variables considered in the distance function.
Each component takes values in $\mathbbm{R}$, initially within the range $[0, 1]$, ensuring that the sum of all components equals 1.
It is the importance of each variable in the distance function.
Genetic algorithms operate on a population of solutions rather than on a single solution.

Each individual is evaluated using a specific fitness function.
The algorithm goal is to find solutions that maximize this function.
Thus, the fitness function has to be defined based on the problem to be solved.
In the customer segmentation problem the fitness function is derived from the objective of the clustering algorithm. It has been defined as the inverse of \kmeans Davies-Bouldin score obtained after applying the \kmeans algorithm. This value quantifies how well the assigned weights allow for a good clustering of the data.

In the initialization stage, an initial population of individuals is generated.
One of the solutions in this initial population is coded out of the weight vectors derived from the relevance values identified during the predictive models built within the EDA process.
The rest are generated randomly.

During each iteration, the quality of each individual in the solution population is assessed using the fitness function.
Then, the best individuals are selected, which have a higher probability of being chosen for breeding.
Selection is performed using a tournament method where several individuals are randomly chosen from the population and compete in a tournament.
The individuals with the best fitness among the selected are chosen for reproduction and to form part of the next generation.
Genetic operators ---selection, crossover, and mutation--- are applied to the individuals.
The mutation operator, specifically a Gaussian mutation, introduces random noise to the weight values, allowing the population to probabilistically explore new solutions.
The outcome of the application of genetic operators on the selected individual is a new population.
An improvement in the general fitness of the population is expected.

This process is repeated until a stopping condition is met, typically when a maximum number of iterations have been executed.
The implemented algorithm model is elitist, ensuring that the best individual from the population is always preserved.

This step of the segmentation procedure has been implemented using the \texttt{DEAP} library.

\subsection{Clustering algorithm}
\label{sec:clustering-algorithm}

The definition of the clustering algorithm is a critical step in our customer segmentation strategy.
This process involves selecting and configuring the appropriate algorithm to effectively group the customer dataset into meaningful segments.
Given the complexities and specific requirements of the dataset, the choice of the right clustering algorithm is crucial for achieving insightful and actionable results.

For this project, we chose the \kmeans clustering algorithm due to its proven efficiency and effectiveness in managing large datasets such as ours.
\kmeans is renowned for its straightforward implementation and rapid convergence, making it a popular choice for a variety of data segmentation tasks.
However, the diverse nature of our data, characterized by variations in scale, density, and type of variables, needed a more customized approach than what standard \kmeans implementations typically offer.

Usual components that have to be defined for \kmeans application are the distance function and the $k$ value. In addition, a stopping criterion must be set.
A further, tweaking point is adjusting the objective function to be optimized during clustering.
All of them have been considered and adjusted in our procedure.
The implemented decisions have been detailed in the preceding sections and are summarized next.


The function definition has been addressed in two broad steps, namely, data transformation (see section~\ref{sec:preprocessing}) and component weight optimization.
Given the scale differences among variables in the dataset, which can greatly affect the performance of \kmeans—since the algorithm uses Euclidean distance as its metric, we normalized the data to ensure that each variable contributes equally to the distance calculations.
This normalization prevents any single feature with a broader range from disproportionately influencing the algorithm behavior, thus ensuring a more balanced and fair clustering process.
Next, we introduced weighted distance metric (see section~\ref{sec:distance-definition}) into the \kmeans algorithm to better handle the variations in the importance and relevance of different variables for clustering.
By assigning weights to different features according to their significance ---determined by EDA and posterior optimization through a genetic algorithms (section{\ref{sec:ga-distance-optimization})--- we were able to finely tune the influence each variable has on the clustering process.
This step is particularly crucial for our dataset, where not all variables are equally informative for the segmentation.
By adapting the distance metric in this way, the \kmeans algorithm can more effectively discern and detect the patterns that are truly significant in the dataset.

Then, optimal $k$ value was found combining DBSCAN with a local search (section~\ref{sec:determining-number-clusters}).
Finally, for selecting the objective function for the clustering process, tests were conducted using inertia \cite{lloyd1982least} and the Davies-Bouldin index \cite{davies1979cluster}.
In our experiments, we observed significant improvements when optimizing the Davies-Bouldin index objective function.
This enhancement is attributed to the index’s ability to generate more distinct and cohesive clusters, making them easier to interpret.
By using this metric clearer and more distinct segmentation of data can be achieved, what facilitates pattern identification and cluster labeling.

We used the \kmeans implementation provided by the \texttt{sklearn} package for this step.

\bigskip
This completes the description of the segmentation methodology designed for our case study. It is convenient to remind that steps 1 to 2 of the procedure are applied once. However, steps 3 to 7 are applied to each partition of the data space. Thus an independent segmentation process is run for on each disjoint partition.

\section{Experimental evaluation}

Once the segmentation methodology had beed defined, it was applied to the dataset.
The specific results attained as well as a thorough analysis are discussed in this section.

The primary target of our experimental process is to optimize the clustering of customer data into meaningful segments. 
As previously detailed in section \ref{sec:dataset-partition}, the process involves a the division of the dataset into distinct segments for analysis by data availability, with each region treated as a separate entity.
This division into regions, which range in size from a few thousand to approximately 1.5 million customer data points, allows for tailored optimization that acknowledges the unique characteristics and challenges presented by each subset of the dataset.

The segmentation procedure, detailed in section~\ref{sec:segmentation-procedure} is almost completely specified. The hyperparameters left to be defined are those related to the genetic algorithm.
The selection process of the genetic algorithm is based on a tournament style, denoted as \texttt{selTournament}, where individuals are grouped into small sets of three (\texttt{tournsize=3}), and winners are determined based on their fitness levels.
Following selection, the genetic operators used are crossover (\texttt{cxBlend}) and mutation (\texttt{mutGaussian}).
The cxBlend crossover blends the features of parent individuals with an $\alpha$ value of 0.5 to create offspring. that combine traits from both parents, potentially leading to better-adapted individuals. It is applied with a crossing probability of 0.7. 
The \texttt{mutGaussian} mutation introduces variability into the population by applying Gaussian noise with a mean ($\mu$) of 0 and a standard deviation ($\sigma$) of 0.2 to the individual’s features, with an individual gene mutation probability (\texttt{indpb}) of 0.2.
This helps prevent the algorithm from settling into local optima and encourages exploration of new areas in the solution space.
The population size is set to 52 individuals, and the algorithm runs for a maximum of 30 iterations (generations).
Elitism is applied meaning that the best individual is preserved from one generation to the next.

Finally, for the \kmeans algorithm we utilize a standard implementation from the \texttt{sklearn} library using 100 estimators and a maximum depth of 300 iterations. 


The optimization of weights using the genetic algorithm significantly improved the fitness in all regions. As an example, a graphical representation of the fitness function evolution in region 5 is plotted in Fig. \ref{fig:fitness}. The genetic algorithm efficiently explored the solution space, adjusting weights to better align with the segmentation goals. This improvement is consistent across all regions, demonstrating the robustness of the approach. The fitness values, which measure the quality of the clustering, showed a marked rise, indicating that the genetic algorithm successfully identified increasingly improved weight configurations for the clustered data.

\begin{figure}[H]
    \begin{center}
    \adjustimage{max size={0.5\linewidth}{0.9\paperheight}}{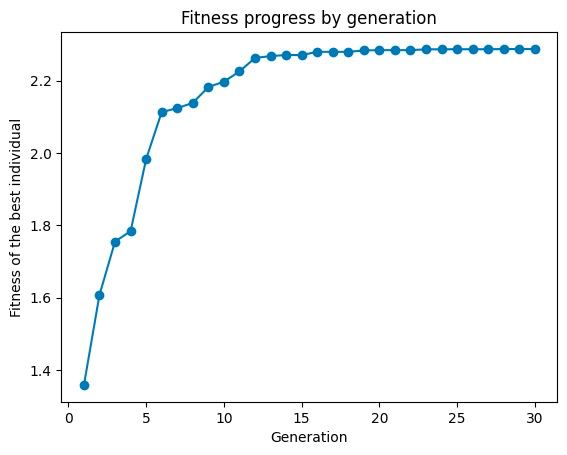}
    \end{center}
    \caption{Graph describing the progress of fitness value over generations during the process of genetic optimization for region 5.}
    \label{fig:fitness}
\end{figure}

As described, the segmentation process derived into eight independent clustering processes.
The overall result is summarized in Table~\ref{tab:clustering-results}.
Since regions are disjoint, the segmentation for all the population is simply the union of the partial clustering. Thus 42 customer profiles have been identified.
Although, initially, this number may seem too high there are two factors to consider. First, is the sheer size of the dataset (close to 3.5M). Second, some regions are rather small, specially regions 3 and 8, that amount to 0.03\% and 0.57\% of the population, respectively.
Removing their groups leave the number of segments in 33. Additionally, some groups in other regions correspond to a very small fraction of the dataset. For example, there are seven groups representing less than 1\% of population.
Removing them leads to a total of 26 relevant segments. 
All the discarded groups altogether cover less than 4.5\% of population.
The reduction in the number of profiles from 42 to 26 allows for a more targeted marketing strategy, concentrating efforts on the most significant segments.
This refinement not only simplifies the campaign management but also enhances the potential impact of marketing initiatives by addressing the needs and preferences of the most prevalent customer groups.

\begin{table}[h]
\centering
\caption{\label{tab:clustering-results}Distribution of clusters by region}
\begin{tabular}{crrr}
\toprule
Region & Cluster & Percentage of the region & Percentage of the global \\
\midrule
1 & 1 & 9.28\% & 1.69\% \\
1 & 2 & 16.17\% & 2.94\% \\
1 & 3 & 28.36\% & 5.16\% \\
1 & 4 & 19.80\% & 3.61\% \\
1 & 5 & 15.71\% & 2.86\% \\
1 & 6 & 10.68\% & 1.94\% \\
\hline 
2 & 1 & 26.71\% & 0.57\% \\
2 & 2 & 23.76\% & 0.51\% \\
2 & 3 & 14.82\% & 0.32\% \\
2 & 4 & 20.10\% & 0.43\% \\
2 & 5 & 14.60\% & 0.31\% \\
\hline
3 & 1 & 19.41\% & 0.01\% \\
3 & 2 & 32.60\% & 0.01\% \\
3 & 3 & 26.19\% & 0.01\% \\
3 & 4 & 21.79\% & 0.01\% \\
\hline 
4 & 1 & 22.79\% & 3.70\% \\
4 & 2 & 37.28\% & 6.05\% \\
4 & 3 & 14.61\% & 2.37\% \\
4 & 4 & 19.84\% & 3.22\% \\
4 & 5 & 5.48\% & 0.89\% \\
\hline 
5 & 1 & 25.84\% & 2.74\% \\
5 & 2 & 20.29\% & 2.15\% \\
5 & 3 & 16.75\% & 1.78\% \\
5 & 4 & 12.77\% & 1.35\% \\
5 & 5 & 10.82\% & 1.15\% \\
5 & 6 & 13.53\% & 1.43\% \\
\hline
6 & 1 & 30.73\% & 3.64\% \\
6 & 2 & 9.31\% & 1.10\% \\
6 & 3 & 16.45\% & 1.95\% \\
6 & 4 & 21.23\% & 2.52\% \\
6 & 5 & 15.56\% & 1.85\% \\
6 & 6 & 6.70\% & 0.79\% \\
\hline
7 & 1 & 20.63\% & 8.34\% \\
7 & 2 & 25.70\% & 10.39\% \\
7 & 3 & 20.91\% & 8.45\% \\
7 & 4 & 22.62\% & 9.14\% \\
7 & 5 & 10.13\% & 4.10\% \\
\hline
8 & 1 & 23.25\% & 0.13\% \\
8 & 2 & 21.90\% & 0.12\% \\
8 & 3 & 24.17\% & 0.14\% \\
8 & 4 & 17.12\% & 0.10\% \\
8 & 5 & 13.56\% & 0.08\% \\
\bottomrule
\end{tabular}
\end{table}

 
\section{Conclusions}

We have addressed the problem of customer segmentation in the sports center business field.
This is a specific case for which a tailored methodology has been designed, implemented and tested.
The peculiarities of the case include a dataset with close to 3.5M instances and severe issues of data quality including high rates of missing values and high presence of outliers, skewed distributions and invalid data values.

The innovation in the proposed methodology falls into data partitioning based on data availability, variable importance analysis through related predictive models, specific distance function definition and optimization based on a genetic algorithm, and \kmeans tweaking.
The application of the methodology to the case resulting in an initial set of 42 groups that could be reduced to 26 relevant ones.

The whole process is detailed and the conclusiones gained can be extended and applied to similar cases.
These comprehensive data handling strategies not only enhance the datasets reliability for segmentation analysis but also support the operational efficiency and marketing strategies of sports centers, ultimately improving the customer experience.

\medskip

\textbf{Acknowledgements}

This research has been funded by Trainingym (Intelinova Software) under a research contract. Trainingym acknowledges fund support by Spanish Government under Grant Call ``2021 Proposals for Research and Development Projects in Artificial Intelligence and other Digital Technologies and their Integration into Value Chains,'' Reference: 2021/C005/0152678.

\bibliographystyle{plain}
\bibliography{refs}

\end{document}